\documentclass{sig-alternate-05-2015}

\usepackage{longtable}

\usepackage{graphicx}
\usepackage{amssymb}

\usepackage{bm,dsfont}
\usepackage{algorithm, algorithmic}
\usepackage{subfigure}
\usepackage{comment}
\usepackage{multirow,tabu,chngpage}

\newcommand{\tab}{\hspace*{2em}}
\newcommand{\halftab}{\hspace*{1em}}
\newcommand{\quadtab}{\hspace*{.5em}}

\begin{document}

\setcopyright{acmcopyright}

\doi{10.475/123_4}

\isbn{123-4567-24-567/08/06}

\conferenceinfo{PLDI '13}{June 16--19, 2013, Seattle, WA, USA}

\acmPrice{\$15.00}

%
\conferenceinfo{ 	
Outlier Definition, Detection, and Description On-Demand}{'16 San Francisco, CA, USA}

\title{Dealing with Class Imbalance using Thresholding}

%
%
%
%
%

\numberofauthors{3} 
%
\author{
%
%
\alignauthor
Charmgil Hong\\
       \affaddr{University of Pittsburgh}\\
       \affaddr{Pittsburgh, PA 15260}\\
       \email{charmgil@cs.pitt.edu}
       \and
\alignauthor
Rumi Ghosh\\
       \affaddr{Robert Bosch LLC}\\
       \affaddr{Palo Alto, CA 94304}\\
       \email{Rumi.Ghosh@us.bosch.com}
\and
\alignauthor Soundar Srinivasan\\
       \affaddr{Robert Bosch LLC}\\
       \affaddr{Palo Alto, CA 94304}\\
       \email{Soundar.Srinivasan@us.bosch.com}
}


\maketitle
\begin{abstract}
We propose thresholding as an approach to deal with class imbalance. We define the concept of thresholding as a process of determining a decision boundary in the presence of a tunable parameter. The threshold is the maximum value of this tunable parameter where the conditions of a certain decision are satisfied. We show that thresholding is applicable not only for linear classifiers but also for non-linear classifiers. We show that this is the implicit assumption for many approaches to deal with class imbalance in linear classifiers. We then extend this paradigm beyond linear classification and show how non-linear classification can be dealt with under this umbrella framework of thresholding.

The proposed method can be used for outlier detection in many real-life scenarios like in manufacturing.  In advanced manufacturing units, where the manufacturing process has matured over time, the number of instances (or parts)  of the product that need to be rejected (based on a strict regime of quality tests) becomes relatively rare and are defined as outliers.  How to detect these rare parts or outliers beforehand? How to detect combination of conditions leading to these outliers? These are the questions motivating our research. This paper focuses on prediction of outliers and conditions leading to outliers using classification. We address the problem of outlier detection using classification. The classes are good parts (those  passing the quality tests) and bad parts (those failing the quality tests and can be considered as outliers). The rarity of outliers transforms this problem into a class-imbalanced classification problem.
\end{abstract}
\printccsdesc



\section{Introduction}
\label{sec:intro}

Consider a manufacturing plant where parts produced are categorized  into  good parts  and bad  parts (scrap) based  on quality tests at the end of an assembly line (EOL tests).
If the scrap rate is low and the quality distribution of the parts is highly skewed (as one can reasonably expect), then the scrapped parts can be defined as outliers.  
If the objective is \emph{scrap detection}, \textit{i.e.}, to predict the scrapped parts before the EOL tests and determine conditions that lead to a scrap, we could transform this problem into a binary classification problem: The two classes would be the predominant good parts and the outlying scrapped parts.

Standard learning algorithms are designed to yield classifiers that maximize accuracy (minimize loss) with an assumption that the class distribution in the dataset is balanced, or nearly-balanced. 
In terms of binary classification, this would imply that the number of observations in one class is roughly equivalent to that of the other class. 
However, this assumption does not hold in the scrap detection problem.
It has been shown that in such scenarios, where the underlying class distribution of data is imbalanced, the standard learning methods cannot perform properly  \cite{He:2009}.   


To further elaborate the issue and outcomes related to the class imbalance, consider the following example. 
If a dataset has $1\%$ observations in the positive class  and $99\%$ observations in the negative class, a classifier that is simply maximizing overall accuracy might put all the observations in the negative class and record $99\%$ accuracy. 
However, for the manufacturing use case, what we are actually interested in is correctly identifying the observations in the positive class. That is, if there are $99\%$ of good parts and $1\%$ of outlying bad parts, the above $99\%$ overall  accuracy would rather be counterproductive and not actionable, and the produced model does not shed much knowledge to detect the outliers.


This problem surfaces frequently not only in manufacturing but also in many other real-world applications.
In \textit{network intrusion} or \textit{fraud detection} \cite{Cieslak:2006:IEEE,Martino:2012:ICPRAM,Phua:2004:KDD}, a very small portion of transactions would be identified as malicious, while almost every transaction is legitimate. 
In \textit{medical diagnosis} \cite{Li:2010:CBM}, predominant patient records indicate the patients are normal, whereas only few patients carry a certain disease.


We propose \emph{thresholding} as a method to deal with the class imbalance problem in classification.  This umbrella framework is defined in terms of a tunable parameter $\alpha$ and a threshold $\alpha^{*}$.The threshold $\alpha^{*}$ is the maximum value for the tunable parameter $\alpha$ wherein the decision choice $d$ is satisfied.  In other words, decision $D=d $ if $\alpha\le \alpha^{*}$. A suitable choice of $\alpha^{*}$ enables one to get actionable insights using classification in the presence of class imbalance.

We present here an illustration of thresholding in the context of binary linear classification, where the classes are labeled by $0$ and $1$ (\emph{i.e}, $D\in \{0,1\}$). The predicted value (output of a classifier) for a class variable of an instance $\alpha$ is often given as a real number between $0$ and $1$ (\emph{i.e.}, $\alpha \in [0,1]$). 
A threshold $\alpha^{*}$ is then determined between $0$ and $1$ such that if the predicted value is less than the threshold, then the instance is predicted to belong to class $0$ (\emph{i.e.}, $D = 0$ \emph{if} $\alpha \le \alpha^{*}$). 
Usually, the threshold is arbitrarily chosen as $0.5$ \emph{i.e.} $\alpha^{*}=0.5$. In this work, we provide a more principled approach to chose the threshold. We show that the ideal choice of the threshold is tightly coupled with the distribution of classes.

We are motivated by the traditional remedies, namely cost-sensitive learning \cite{Elkan:2001:IJCAI,Weiss:2004:SIGKDD,Zhou:2006:IEEE,Liu:2006:ICDM} and sampling techniques \cite{Japkowicz:2002:IDA,ling:1998:kdd,Batista:2004:SIGKDD,Elkan:2001:IJCAI}, that adjust the decision threshold to increase true positives (TP; correct predictions on the minority class instances). However, an application of such remedies often tends to  overcompensate for true positives (TP)  by sacrificing true negatives (TN). Though the right compromise is often difficult to reach, it is of paramount importance to  have a guiding stick to reach an acceptable trade-off. Taking the manufacturing use case where the positives are the scrapped parts and the negatives are the good parts,  it is  unfavorable to  have a low precision ($\frac{TP}{TP+FP}$) because the false alarms (FP) may incur expensive follow-up actions.  Our thresholding approach  provides guidance in the direction of attaining the optimal trade-off. 

Our contributions are three-fold:
\vspace{-1em}
\begin{itemize}
\item Firstly, we formalize the concept of thresholding  and provide a novel perspective to classification using the concept as an umbrella framework. We show that the method of thresholding can be used to address class imbalance both for linear and non-linear classification. 
\vspace{-2em}
\item Secondly, for linear classifiers when the observed classes are discrete and the prediction is a real value, we provide a principled approach for choosing the threshold for the real-valued prediction to decide the predicted class of the observation. This threshold is based on the distribution of the classes in the dataset. This technique enables classification even in severely imbalanced datasets. If the class with fewer instances comprises of the outliers, this enables outlier detection.
\vspace{-1em}
\item Thirdly, we provide a novel method of thresholding for non-linear classifiers like decision trees.  In decision trees, we use the divide-and-conquer approach, which can define separate regions in the input space and learn a distinct threshold on each region.  In particular, we propose a new method to define such subregions using the R{\'e}nyi entropy \cite{renyi:1961}. We study the relations between the entropy and the class imbalance ratio in a subregion, and develop an algorithm to exploit the relationship for decision tree learning.
\end{itemize}

We begin with a  review of  related research (Section \ref{sec:related}). Next, we deal with the concept of thresholding in linear classifiers (Section \ref{sec:linear}). We show that this is an implicit assumption for many approaches to deal with class imbalance. We then extend this paradigm beyond linear classification and show how decision trees can be dealt with under this umbrella framework of thresholding (Section \ref{sec:dt}).
Lastly we present experimental results that show the accuracy and robustness of our proposed method (Section \ref{sec:exp}) and conclusion.

We would like to note that our choice of the algorithm subset, which is investigated under the  proposed umbrella framework of thresholding in this paper, is motivated by interpretability. Taking the scrap detection use case in our introduction, it has been observed that the interpretable classification techniques achieve greater buy-in in non-traditional data mining domains like manufacturing.

\section{Related Research}
\label{sec:related}

The class imbalance problem has been extensively studied in the literature. In this section, we review some of the representative work that are closely related to our work.

\textbf{Sampling} is arguably the simplest and the most widely used approach to deal with the class imbalance problem. The main idea is to rebalance the dataset such that the standard classification method can effectively fit the data without algorithmic modifications. Depending on how the sampling is done, the approach can be categorized as: \textit{Random under-sampling} under-samples the majority class instances \cite{Elkan:2001:IJCAI,Japkowicz:2002:IDA}; \textit{random over-sampling} over-samples the minority class instances \cite{ling:1998:kdd,Batista:2004:SIGKDD}; and \textit{synthetic data injection} generates new synthetic samples according to the minority class distribution \cite{Chawla:2002:JAIR,Cohen:2006:AIM}.

Another widely accepted approach is \textbf{cost-sensitive learning} \cite{Elkan:2001:IJCAI}. This approach tackles the class imbalance problem by exploiting the cost matrix that defines the costs (penalties) associated with TP, FP, TN, and FN \cite{Weiss:2004:SIGKDD,Zhou:2006:IEEE,Liu:2006:ICDM}. In particular, a misclassification of a minority class instance (FN) involves higher cost than that of a majority class instance (FP); whereas correct classifications (TP and TN) typically do not involve costs. By minimizing the classification cost (Equation \ref{eq:4}), one can train a classifier that takes the class imbalance into account.

On the other hand, the \textbf{decision tree approaches} have been very different from the former two approaches. 
The idea is to modify the splitting criteria such that the decision tree learning algorithm can discover useful decision branches and, hence, build effective decision trees even in the presence of class imbalance. 
\cite{Dietterich:1996:ICML} proposed a splitting criterion in an effort to obtain more robust decision trees. Although its original objective was to improve the learning algorithm to satisfy the PAC learning condition \cite{Valiant:1984:TL}, later the proposed criterion was shown to improve the predictive accuracy of the decision trees on imbalanced datasets \cite{Drummond:2000:ICML}.
\cite{Liu:2010:SDM} and \cite{Cieslak:2012:DMKD} further studied the relationship between the splitting criterion and the class distribution. In particular, they investigated the effect of the underlying class imbalance on different impurity measures and proposed new decision tree learning algorithms that use the class confidence proportion \cite{Liu:2010:SDM} and the Hellinger distance \cite{Cieslak:2012:DMKD} as the splitting criterion.

The R{\'e}nyi entropy \cite{renyi:1961} has been applied to decision tree learning as an effort to obtain effective decision models from imbalanced data. \cite{Maszczyk:2008} and \cite{Lima:2010} simply used the R{\'e}nyi entropy as a substitute of the Shannon entropy and showed that the R{\'e}nyi entropy can be useful in learning a robust decision tree on imbalanced data, given a proper choice of the parameter $\alpha$ (which is fixed throughout the learning). However, proper parameter choices are not known a priori and, hence, one has to run with multiple parameter candidates to find the best among them. Later, \cite{Park:2014} attempted to alleviate the issue by proposing ensembles of $\alpha$-trees. That is, they used the R{\'e}nyi entropy with multiple parameters to obtain diverse trees (each tree is trained with a fixed $\alpha$) from data for building ensemble models. 
However, the ensemble decision is made by a simple majority vote which does not show consistent results in practice (see Section \ref{sec:exp}). 

In this work, we study the concept of thresholding as a general imbalance-sensitive model improvement approach. 
Our approach incorporates thresholding with decision tree learning by devising a new splitting criterion that changes adaptively according to the underlying class distribution. Although we adopt the same R{\'e}nyi entropy as the above mentioned methods, our method is different in that it decides the parameter $\alpha$ according to the class distribution at each decision node and, as a result, provides more accurate and stable performance.

\section{Addressing Class Imbalance with Linear Classifiers}
\label{sec:linear}
 In this section, we  define a class of linear models and show how to adjust their decision threshold to fit the underlying class distribution in data.
We then briefly overview two of the widely used methods that address the class imbalance problem in the context of learning linear classifiers~--\emph{ cost-sensitive learning} and \emph{sampling}~-- and relate these methods using an umbrella concept of \textit{thresholding}. 

\subsection{A Class of Linear Models}

This section defines a class of statistical models that generalizes linear regression and logistic regression. 
Let $\mathcal{X}= \{ \mathbf{x}_i \}_{i=1}^n$ and $\mathcal{Y}= \{ y_i \}_{i=1}^n$ be the variables of our interest, where $\mathbf{x}_{i}$ is a length $m$ feature vector (input) and $y_i$ is its associate output variable.
We refer to $p_i$ as a linear estimator of $y_i$, if it is of the form $p_i=g(\sum_{j=1}^{m} w_{j}x_{ij}+ w_{0})$.
Below we provide a generalized theorem which shows that \textit{$p_i$ varies linearly with the class imbalance for these linear classifiers}. 
\newtheorem{theorem}{Theorem}
\begin{theorem}
Let $\theta_i= \sum_{j=1}^{m} w_{j}x_{ij}+ w_{0}$. Given that the observed variable of interest is $y_i$, we denote its linear estimator by $p_i=g(\theta_i)$.
For all linear classification function of the form: 
\begin{equation}
L(\mathcal{X},\mathcal{Y})=\sum_{i=1}^{n} q(\theta_i)- y_i\theta_i +c 
\label{eq:3}
\end{equation}
 where $\frac{dq}{d\theta}=g$
and $c$ is a constant, the estimated value of the variable of interest varies linearly with the ratio of class imbalance.
\label{th:1}
\end{theorem}
\vspace{-1em}
\begin{proof}
By differentiating with respect to ${w_{j}}$, we get:
$\frac{dL(\mathcal{X},\mathcal{Y})}{dw_{j}}=\sum_{i=1}^{n}( \frac{dq}{d\theta}x_{ij}-y_i x_{ij} )=\sum_{i=1}^{n}( g(\theta_i)x_{ij}- y_i x_{ij})$.
By differentiating with respect to $w_0$, we get:\\
 $\frac{dL(\mathcal{X},\mathcal{Y})}{dw_{0}}=\sum_{i=1}^{n}( g(\theta_i)- y_i )=\sum_{i=1}^{n}( p_i-y_i)$.
To minimize this loss function, taking  $\frac{dL(\mathcal{X},\mathcal{Y})}{dw_{0}}=0$ gives us:
\begin{equation}
\sum_{i=1}^{n} y_i =\sum_{i=1}^{n} p_i
\label{eq:1}
\end{equation}
We suppose that observations  $y_\nu \in \{0, 1\}$ are drawn from populations having exponential power distribution with means  $\overline{Y}_{\nu} \in \{ \overline{Y}_{0}, \overline{Y}_{1} \}$, respectively. Assuming that the samples are sufficiently large and taking $\widehat{Y}_{\nu}$ as the sample means, we have $\overline{Y}_{\nu} \approx \widehat{Y}_{\nu}$.
If the ratio of the binary classes $1$ and $0$ is $\mu:1-\mu$, then depending on the class imbalance Equation \ref{eq:1} can be rewritten as:
\begin{align}
n(\mu \overline{Y}_{1}+(1-\mu) \overline{Y}_{0}  ) =\sum_{i=1}^{n} p_i= n \overline{p}
\label{eq:2}
\end{align}
Here $\overline p$ is the sample mean of the linear estimator. Notice that this sample mean varies linearly with the ratio of class imbalance. 
\end{proof}

\newtheorem{lemma}{lemma}

\begin{lemma}
When the linear estimator is a logistic regressor, \textit{i.e.}, $g(\theta_i)=\frac{1}{1+e^{-\theta_i}}$, then {Theorem} \ref{th:1} implies that an appropriate loss function to minimize would be the log-likelihood loss function.
\end{lemma}

\subsection{Direct Approach of Thresholding with Linear Classifiers}
\label{subsec:linear}

Without loss of generality, we continue to deal with the linear binary classifiers that form $p_i=g(\sum_{j=1}^{m} w_{j}x_{ij} + w_{0})$. 
Let us further assume that $p_i \in [0,1]$ and $p_i$ can be interpreted as the estimated probability that $y_i=1$ on the $i$-th observation $x_i$.
%
In this section, we show that the negative effect of class imbalance to the linear classifiers, can be alleviated by adjusting the decision threshold.
By rewriting Equation \ref{eq:3} as the summations over the two classes, we obtain:  
\begin{align}
&L(\mathcal{X},\mathcal{Y})=\notag\\
&\quadtab\sum_{i \in \{y_i = 1\}} (q(\theta_i)- y_i\theta_i +c ) + \sum_{i \in \{y_i = 0\}} (q(\theta_i)- y_i\theta_i +c )
\label{eq:3-a}
\end{align}
%
We let $\overline p_{\{y_i = \nu\}} (\nu \in \{0,1\})$ represent the population mean of class $\nu$, and $\mu$ denote the class imbalance ratio. By minimizing Equation \ref{eq:3-a} with respect to $w_0$, we obtain:
\begin{equation}
n(\mu \overline Y_{1}+(1-\mu) \overline Y_{0}  ) =n(\mu \overline p_{\{y_i = 1\}}+(1-\mu)\overline p_{\{y_i = 0\}})
\end{equation}
As in Equation \ref{eq:2}, $\overline Y_{\nu}$ denotes the mean values of observations for the populations of class $\nu$.
Now, knowing $\overline Y_\nu = \nu$ gives us:
\begin{equation}
\mu \overline p_{\{y_i = 1\}}+(1-\mu)\overline p_{\{y_i = 0\}}=\mu
\label{eqi:5}
\end{equation}

Let $\alpha^{*}$ be the threshold such that if $p_i<\alpha^{*}$, then $x_i$ is classified as $y_i = 0$; otherwise, it is classified as $y_i = 1$. 
If $p_i=\alpha^{*}$ then it has equal probability of belonging to class $0$ or class $1$. 
In other words, the \emph{normalized} distances from the mean should be equivalent when $p_i=\alpha^{*}$; 
\textit{i.e.}, $\frac{\alpha^{*}- (1-p_i)}{1-\mu}=\frac{p_i - \alpha^{*}}{\mu}$. 
This implies:
\begin{equation}
\mu p_i+(1-\mu)(1-p_i)=\alpha^{*}
\label{eqi:6}
\end{equation}

From Equations \ref{eqi:5} and \ref{eqi:6}, we get $\alpha^{*}=\mu$, \textit{i.e.}, there is a direct mapping between threshold and imbalance.
A point $p_i$ belongs to class $1$ if  $\frac{p_i- \overline p_{\{y_i = 0\}}}{1-\mu}>\frac{\overline p_{\{y_i = 1\}}- p_i}{\mu}$; \textit{i.e.}, $p_i>\alpha^{*}$.
That is, given imbalance $\mu$, we can decide the threshold such that the classification model takes the imbalance into account.



\subsection{Indirect Approach of Thresholding with Linear Classifiers}
\label{subsec:balancing}

In the previous section, we discussed a direct approach to the class imbalance problem with the linear classifiers which is essentially to shift the decision threshold along with the imbalance ratio.
This section describes an indirect approach to cope with the imbalance problem in linear classification.

Recall that most standard classifiers implicitly assume that the dataset is balanced and, hence, often the decision threshold is $0.5$, \textit{i.e.}, if $p_i\le0.5$ then $y_i=0$ and $1$ otherwise. 
When the dataset is balanced, linear estimators $\overline Y_{1}, \overline Y_{0}, p_i$ and $\overline p$ satisfy the following equation:
\begin{equation}
\frac{n}{2}(\overline Y_{1}+ \overline Y_{0}  ) =\sum_{i=1}^{n} p_i= n \overline p
\label{eq:7}
\end{equation}
The indirect approach works by rebalancing an imbalanced dataset such that the resultant estimator $p_i$ becomes (roughly) balanced
and, therefore, the standard learning algorithms can perform reasonably well without making fundamental changes to the model.
More specifically, the approach adjusts the importance that is associated with each class such that the positive class instances contribute more towards model learning. 
Conventionally, such a rebalancing is achieved by either the cost-sensitive approach \cite{Elkan:2001:IJCAI} or the sampling techniques.

\vspace{.5em}
\noindent
\textbf{Cost-Sensitive Learning:} \quadtab One way to rebalance data is to increase the importance associated with the misclassification with the rarer or the outlier class, as opposed to associating the same importance to all misclassification. This leads to the cost-sensitive learning approach \cite{Elkan:2001:IJCAI}. Recall that as the class imbalance ratio $\mu$ increases, the expected value $E[p]$ becomes more biased towards the sample mean of class $1$ (Equation \ref{eq:2}).  If $\mu\gg1-\mu$ then class $1$ would be the predominant class. In the cost-sensitive approach, we associate distinct costs $c_0$ and $c_1$ respectively  with class 0 and 1. Equation \ref{eq:3} can be extended for the cost-sensitive learning as:
\begin{align}
&L(\mathcal{X},\mathcal{Y}) =\notag\\ 
&\halftab\quadtab c_1(\sum_{i=1}^{n \mu} q(\theta_i)- y_i\theta_i +c ) + c_0(\sum_{i=1}^{n (1-\mu)} q(\theta_i)- y_i\theta_i +c )
\label{eq:4}
\end{align}
Equation \ref{eq:2} is then generalized to: 
\begin{align}
c_1 n \mu \overline Y_{1}+c_0 n (1-\mu) \overline Y_{0} =c_1\sum_{i=1}^{n \mu} p_i +c_0\sum_{i=1}^{n (1-\mu)} p_i \nonumber
\end{align}
Taking $c_1= \frac{1}{2  \mu  }$ and  $c_0= \frac{1}{2 (1- \mu)  }$ leads to:
\begin{align}
\frac{n}{2}(\overline Y_{1}+\overline Y_{0}) =\frac{1}{2 \mu  }\sum_{i=1}^{n\mu} p_i +\frac{1}{2 (1- \mu)  }\sum_{i=1}^{n(1-\mu)} p_i=n \bar p \nonumber
\end{align}
The expected value of the linear estimator obtained in Equation \ref{eq:7} is equivalent to that obtained when the two classes are balanced in the dataset.

\vspace{.5em}
\noindent
\textbf{Sampling:} \quadtab
We note that introducing costs while dealing with imbalanced datasets leads to the change in loss function from Equation \ref{eq:3} to Equation \ref{eq:4}.
One way to introduce cost-sensitivity into decision making without changing the loss function is using sampling techniques. Having different sampling frequency for the two classes enables us to use the algorithms designed for balanced datasets for imbalanced datasets. 


\section{Addressing Class Imbalance with Decision Trees}
\label{sec:dt}

In the previous section, we reviewed how to bias the decision threshold of linear classifiers and adjust them according to the imbalance in data. In this section, we extend the concept of thresholding towards decision trees and propose a novel decision tree learning algorithm, called Adaptive R{\'e}nyi Decision Tree (ARDT).

\subsection{Standard Decision Trees}
\label{subsec:standard-dt}

Quinlan \cite{Quinlan:1987,Quinlan:1993} has introduced a decision tree learning algorithm that recursively builds a classifier in a top-down, divide-and-conquer manner. Due to its simplicity and computational efficiency, the algorithm has been widely accepted and extended in various ways \cite{Kotsiantis:2013:AIR}. 
On a training dataset $\mathcal{D} = \{ \mathbf{x}_i,y_i \}_{i=1}^n$, the algorithm learns a decision tree as below.
\\

\vspace{-1em}
\begin{algorithmic}[1]
    \STATE Select the best splitting feature and value on $\mathcal{D}$ according to a \textit{splitting criterion}.
    \STATE Create a decision node that splits on the feature and value selected; correspondingly, partition $\mathcal{D}$ into $\mathcal{D}_L$ and $\mathcal{D}_R$.
    \STATE Repeat steps {\small 1-2} on $\mathcal{D}_L$ and $\mathcal{D}_R$ until all the leaf nodes satisfy \textit{stopping criteria}.
\end{algorithmic}

On each recursion, the tree grows by turning a leaf node into a decision split which has two or more child nodes (the above algorithm illustrates a binary split only for simplicity). 
The tree stops growing when all leaf nodes suffice certain \textit{stopping criteria} which are usually defined by a combination of conditions, such as purity rates, node sizes, and tree depth. 

How to split a leaf node is determined by a \textit{splitting criterion}, which measures the impurity in a collection of training instances. The most commonly used splitting criterion is \textit{information gain} (IG) based on the Shannon entropy \cite{shannon:1948}:
\begin{align}
\label{eq:infogain}
&\textit{IG} =  H_\textit{Shannon}(Y) - E_x\left[ H_\textit{Shannon}(Y | x) \right],
\end{align}
where
\begin{align}
&H_\textit{Shannon}(Y) = -\sum_{y\in Y} P(y) \log_2 P(y)
\label{eq:shannon}
\end{align}
IG measures the expected reduction in entropy after the split specified by $x$.
Equation \ref{eq:shannon} defines the Shannon entropy. It ranges between 0 and 1: it is maximized when $P(y)$ is uniform, and minimized when $P(y)$ is close to either 0 or 1 (see Figure \ref{fig:entropies}). 
As a result, the Shannon entropy measures the impurity in a data collection and, therefore, we can identify the best split by minimizing the expected entropy (Equation \ref{eq:infogain}).
\\

\noindent
\textbf{Reduced-error Pruning:} \quadtab
The top-down induction of decision trees often results in models overfitted to training data. A common solution to this issue is pruning. To prune a decision tree, we traverse a unpruned tree in post-order (traverse the subtrees first, then the root) and replace a subtree into a leaf node if the replacement does not worsen a \textit{pruning criterion}. \cite{Quinlan:1987} has proposed the \textit{reduced-error} pruning criterion, with which subtree replacements are made according to the overall error $(\frac{FP+FN}{N})$. This criterion has been accepted as a rule of thumb in many application domains to alleviate the overfitting issue of decision trees.

\subsection{Effects of Class Imbalance on Standard Decision Trees}
\label{subsec:standard-dt-imbalance}
Although the information gain criterion based on the Shannon entropy has shown preferable performances in many applications \cite{Kotsiantis:2013:AIR}, the criterion may not work properly when the dataset is imbalanced ($P(y) \ll 0.5$) \cite{Maszczyk:2008,Lima:2010}. That is, when using the criterion on imbalanced data, the produced classifier often becomes biased towards the negative class and ignores the positive class. 
The rationale behind this unfavorable behavior can be found easily using the Bayes' theorem:
\begin{align}
E_x\left[ H_\textit{Shannon}(Y | x) \right]
& = E_y\left[ -\sum_{x} P(x|y) \log_2 P(y|x) \right]\notag
\end{align}
Consequently, the influence of each class $y \in Y$ to the Shannon entropy is proportional to $P(Y=y)$. 
This will further confuse the decision tree learning process and hinder us from obtaining accurate classifiers. 
In the next subsection, we present our approach that subes this undesirable behavior on imbalanced data. 
\\

\noindent
\textbf{Validity of Reduced-error Pruning on Imbalanced Data:} 
When data is imbalanced the \emph{reduced error} pruning criterion may not be satisfactory, because the overall error is often dominated by FP, which in turn results in unwanted neglect on FN. In  our approach in the next subsection, we show how to avoid this negative outcome with a simple modification of the criterion. 


\subsection{Our Approach}
In this subsection, we propose a new decision tree learning method for the class imbalance problem, called Adaptive R{\'e}nyi Decision Tree (ARDT), which applies the thresholding idea to adapt its splitting criterion according to the underlying class distribution at each decision node. 
We then present our pruning criterion that does not bias towards the negative class.

\subsubsection{Learning Decision Trees in Consideration of Class Imbalance}
\label{subsec:learning}

\begin{figure}[t]
	\centering
	\includegraphics[width=0.45\textwidth]{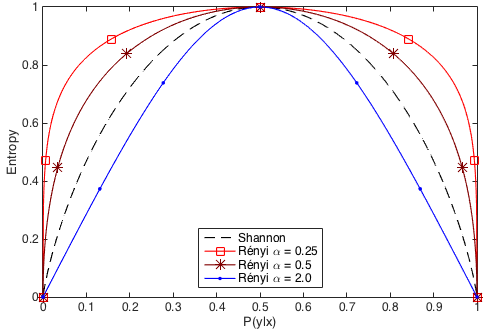}
	\caption{The R{\'e}nyi and Shannon entropy curves.}
	\label{fig:entropies}
\end{figure}


In Section \ref{subsec:standard-dt-imbalance}, we discussed that the Shannon entropy becomes unreliable when the class prior distribution $P(y)$ is highly skewed to one class. Knowing that the conventional entropy measure may get adversely affected by $P(y)$, we develop a method that automatically adjusts the metric according to $P(y)$. In particular, we propose to use the R{\'e}nyi entropy \cite{renyi:1961} as a new splitting criterion. The R{\'e}nyi entropy is defined as: 
\begin{align}
\label{eq:renyi}
H_\textit{R{\'e}nyi} = \frac{1}{1-\alpha} \log \sum_y P(y | x)^\alpha,
\end{align}
where $\alpha$ is a user parameter that determines the operating characteristics of the entropy measure. Figure \ref{fig:entropies} shows how the entropy changes according to $\alpha$. When $\alpha$ decreases from 1 to 0, the region where the entropy is maximized becomes wider; while $\alpha$ increases from 1, the arc shape turns thinner and the region where the entropy is maximized becomes narrower. Note that the R{\'e}nyi entropy generalizes the Shannon entropy. That is, the R{\'e}nyi entropy tends to the Shannon entropy as $\alpha \rightarrow 1$. For a more theoretical review of the R{\'e}nyi entropy, see \cite{vanErven:2014:IEEE}. 

Although the R{\'e}nyi entropy has been applied to decision tree learning in \cite{Maszczyk:2008,Lima:2010}, their extensions are limited in that they simply replaced the entropy measure and hardly exploited the relationship between the different operating characteristics driven by parameter $\alpha$ and the class prior distribution $P(y)$. 
In the following, we study the relationship between $\alpha$ and $P(y)$ and show how we incorporate the concept of thresholding in developing our new learning algorithm.

Without loss of generality, we discuss the decision tree learning process at an arbitrary decision node $l$. Let $P_l(y)$ class denote the prior distribution at node $l$. Let $P_l(y|x < a)$ and $P_l(y|x > a)$ be the distributions on the partitions from node $l$, where $x < a$ and $x > a$ represent a binary partition. Note that $P_l(y|x < a) = P_{2l}(y)$ and $P_l(y|x > a) = P_{2l+1}(y)$ will become the children of node $l$. 
Also note that, on each decision node, the class prior is changing; \textit{i.e.}, $P_l(y) \leq P_{2l}(y)$ and $P_l(y) \geq P_{2l+1}(y)$, or vice versa.
Now recall that on imbalanced data the Shannon entropy may become biased towards the negative class. This can be seen more clearly on the entropy curve: In Figure \ref{fig:entropies}, the dashed line draws the Shannon entropy. When working with a dataset where only few instances fall in class $Y=1$ (and the majority of them fall in $Y=0$), there will be many candidate partitions whose $P(y|x)$ is close to 0. This results in the overestimation of information gain on arbitrary partitions and may lead to a decision tree that favors the negative class instances.

By adopting the R{\'e}nyi entropy, we can alleviate this undesirable behavior by adjusting its parameter $\alpha$ according to the class prior distribution at node $l$, $P_l(y)$. More specifically, we set parameter $\alpha$ to maximize $H_\textit{R{\'e}nyi}$ on $P_l(y)$ and promote purer partitions (\textit{e.g.}, $\min(P_l(y), 1-P_l(y)) > \min(P_l(y|x < a), 1-P_l(y|x < a))$). 
Let $\alpha^*$ be such a value of the parameter. Then, threshold $\alpha^*$ can be found by seeking the largest value of $\alpha$ that satisfies $H_\textit{R{\'e}nyi}(\alpha,P_l(y)) = 1$. This can be formally written as:
\begin{align*}
&\alpha^* = \max\alpha,\\
\text{subject to}\tab&\frac{1}{1-\alpha} \log \sum_{y\in Y} P_l(y)^\alpha=1
\end{align*} 
Assuming the underlying class distribution is continuous and at least twice differentiable, we can analytically derive $\alpha^*$ as:
\begin{align}
\alpha^* = \frac{P_l(y)}{P_l'(y)+P_l(y)} \left( 1 + \frac{P_l'(y)}{P_l(y)} + \frac{P_l''(y)}{P_l'(y)} \right)
\label{eq:findalpha}
\end{align}
However, such assumptions do not always hold when the target variable $Y$ is discrete. Therefore, instead of using Equation \ref{eq:findalpha}, we heuristically find $\alpha$ using a sequential search. Algorithm \ref{alg:findalpha} implements this search procedure. 
By varying $\alpha$ from 1 to 0 (with a step size $\epsilon$), it attempts to find the largest $\alpha$ that satisfies $H_\textit{R{\'e}nyi}(\alpha, P_l(y)) = 1$. 
Notice that we are switching back to the Shannon entropy when $P_l(y)$ is equal to 0.5; that is, when the class prior distribution is balanced. 

Lastly, Algorithm \ref{alg:rbdt-train} summarizes our proposed decision tree learning algorithm.


\begin{algorithm}[t]
    \caption{\textit{Find-alpha}}
    \label{alg:findalpha}
    \mbox{\small{\textbf{Input}: Class prior $P(y)$, Step size $\epsilon$}}\\
    \mbox{\small{\textbf{Output}: R{\'e}nyi parameter $\alpha^*$}}
\algsetup{linenosize=\small}
\small
\begin{algorithmic}[1]
    \FOR{$\alpha = 1, 1-\epsilon, ..., \epsilon, 0$}
        \IF{$H_\alpha(P_l(y)) = 1$}
            \STATE \textbf{Return} $\alpha$
        \ENDIF
    \ENDFOR
\end{algorithmic}
\end{algorithm}

\begin{algorithm}[t]
    \caption{\textit{ARDT-Train}}
    \label{alg:rbdt-train}
    \mbox{\small{\textbf{Input}: Train data $\mathcal{D} = \{\mathbf{x}_i,y_i\}_{i=1}^n$}}\\
    \mbox{\small{\textbf{Output}: Decision tree $T$}}
\algsetup{linenosize=\small}
\small
\begin{algorithmic}[1]
    \STATE Determine $\alpha^*$ using Algorithm \ref{alg:findalpha}
    \STATE Select the best splitting feature and value by Equation \ref{eq:renyi}
    \STATE Using the selected feature and value, create a decision node and partition $\mathcal{D}$ into $\mathcal{D}_L$ and $\mathcal{D}_R$
    \STATE Build subtrees by recursion; \textit{i.e.}, \textit{ARDT-Train}($\mathcal{D}_L$) and \textit{ARDT-Train}($\mathcal{D}_R$)
\end{algorithmic}
\end{algorithm}

\subsubsection{Pruning Decision Trees}
\label{subsubsec:pruning}

In order to avoid the unwanted neglect of the positive class instances through pruning (see Section \ref{subsec:standard-dt}), we prune the tree using the Balanced Classification Rate (BCR) \cite{Lannoy:2011}, which is the geometric mean of sensitivity and specificity, as a pruning criterion:
\begin{align}
BCR 
&=\frac{1}{2}\left(\frac{TP}{TP+FN} + \frac{TN}{TN+FP}\right)
\end{align}
By considering sensitivity and specificity independently, we prevent the negative class dominates the pruning and effectively suppress the misclassification of both classes (FN and FP). We refer our pruning criterion as to \textit{improved-BCR} pruning.

\section{Experiments}
\label{sec:exp}

To demonstrate the effectiveness of the ARDT method, we perform two sets of experiments.
In Section \ref{subsec:exp1}, we evaluate the accuracy of our proposed method and its robustness against class imbalance using publicly available imbalanced datasets. Specifically, we compare the performance of ARDT and other baseline methods, and analyze their predictive accuracy in the presence of class imbalance.
In Section \ref{subsec:exp2}, we apply ARDT to a real world outlier detection problem, where we want to automatically identify bad parts or scrap produced in a manufacturing assembly line. Through the study, we empirically verify the usefulness of ARDT in addressing the problem. 

\subsection{Evaluation with Public Data}
\label{subsec:exp1}


We perform experiments on 18 publicly available imbalanced datasets that are listed in Table \ref{table:datasets}\footnote{All datasets are obtained from \url{https://archive.ics.uci.edu/ml/datasets.html} and \url{http://sci2s.ugr.es/keel/imbalanced.php}}. These datasets are obtained from various domains, including audio categorization (Birds \cite{Briggs:2013}), optical character recognition (Letter \cite{frey:1991}), scientific classification (Glass \cite{evett:1987}, Yeast \cite{Elisseeff:2001}, and Ecoli \cite{Horton:1996}), medical diagnosis (Thyroid \cite{Quinlan:1987:ACAES}), and industrial classification (Pageblock \cite{Esposito:1994} and Led7digit \cite{Breiman:1984}).

\vspace{.5em}
\noindent
\textbf{Methods:} \quadtab
We compare the performance of our proposed method, which we refer to as Adaptive R{\'e}nyi Decision Tree (ARDT), with the conventional techniques treating the class imbalance problem, including the cost-sensitive learning and sampling. More specifically, we compare ARDT with the linear regression (LinR) and logistic regression (LogR) models that are trained with a cost matrix \cite{Weiss:2004:SIGKDD} or are trained on under/over-sampled data \cite{Japkowicz:2002:IDA,Batista:2004:SIGKDD}. As a result, our baselines include the following eight combinations: 
\textit{standard linear regression (LinR)}, \textit{cost-sensitive LinR (LinR+CS)}, \textit{LinR with random under-sampling (LinR+US)}, \textit{LinR with random over-sampling (LinR+OS)}, 
\textit{standard logistic regression (LogR)}, \textit{cost-sensitive LogR (LogR+CS)}, \textit{LogR with random under-sampling (LogR+US)}, and \textit{LogR with random over-sampling (LogR+OS)}.

We also compare ARDT with the standard decision tree \cite{Quinlan:1993} and its variants that are designed to solve the class imbalance problem. These include: \textit{C4.5 (CDT)} \cite{Quinlan:1993}, \textit{Decision trees using DKM (DKMDT)} \cite{Dietterich:1996:ICML}, \textit{Hellinger Distance Decision Tree (HDDT)} \cite{Cieslak:2012:DMKD}, and \textit{Ensemble of $\alpha$-Trees (EAT)} \cite{Park:2014}. For all decision tree models, we prune the decision trees using the \textit{improved-BCR} criterion (see Section \ref{subsubsec:pruning}).

\begin{table}
\begin{center}
    \resizebox{.35\textwidth}{!}{
    \begin{tabular}{lcccc}
    \hline
     Dataset         & $N$    & $m$   & $P(Y=1)$  & Domain\\ \hline
    Birds-s-thrush		& 645  & 276 & 0.16 & Audio  \\
    Birds-kinglet			& 645  & 276 & 0.06 & Audio  \\
    Birds-nighthawk		& 645  & 276 & 0.04 & Audio     \\
    Letter-A				& 20,000  & 16 & 0.04 & Image     \\
    Letter-B				& 20,000  & 16  & 0.04 & Image    \\
    Letter-C			& 20,000 & 16  & 0.04 & Image     \\
    Letter-D			& 20,000 & 16  & 0.04 & Image       \\
    Glass-containers		& 214  & 9   & 0.06 & Chemistry        \\
    Glass-tableware		& 214  & 9   & 0.04 & Chemistry       \\
    Yeast-vac-vs-nuc		& 459  & 7   & 0.07 & Biology    \\
    Yeast-me2			& 1,484  & 8   & 0.10  & Biology    \\
    Yeast-me1			& 1,484 & 8   & 0.03 & Biology    \\
    Yeast-exc			& 1,484 & 8   & 0.11 & Biology    \\
    Ecoli-om			& 336 & 7   & 0.06 & Biology    \\
    Thyroid1			& 215  & 5   & 0.16 & Medical \\
    Thyroid2			& 215  & 5   & 0.16 & Medical \\
    Pageblocks0			& 5,472 & 10  & 0.10  & Industry \\
    Led7digit1			& 443 & 7  & 0.08  & Industry \\ \hline\\
\end{tabular}}
\caption{Dataset characteristics ($N$: number of instances, $m$: number of features, $P(Y=1)$: class imbalance ratio)}
\label{table:datasets}
\end{center}
\end{table}

\vspace{.5em}
\noindent
\textbf{Metrics:} \quadtab
We use the following evaluation metrics to compare the methods.
\vspace{-.5em}
\begin{itemize}
\item F1-score (FSCORE): FSCORE measures the harmonic mean of the precision and sensitivity of a classifier. It provides a reasonable summary of the performance on each of the majority and minority classes, and therefore is of our primary concern.
$$\textit{FSCORE} = \frac{\textit{\small Precision} \cdot \textit{\small Sensitivity}}{\textit{\small Precision} + \textit{\small Sensitivity}}
= \frac{\text{\small 2}\cdot \text{\small $TP$}}{\text{\small 2}\cdot \text{\small $TP + FP + FN$}}$$
\vspace{-1em}
\item Accuracy (ACC): ACC measures how correctly a method classifies instances. Although it may not precisely reflect how a method behaves (\textit{e.g.}, blindly predicting every instance as the majority class could achieve higher ACC), since it is an important metric in many applications, we include it in our discussion. 
\end{itemize}
\vspace{-1em}



\begin{table*}[h!]
    \centering

    \resizebox{1\textwidth}{!}{
    \begin{tabular}{l| c cccc c cccc c ccccc}
    \multicolumn{1}{c}{\multirow{2}{*}{FSCORE}}    & ~ & \multicolumn{4}{c}{\textit{Linear Regression (LinR)}}    & ~ & \multicolumn{4}{c}{\textit{Logistic Regression (LogR)}}    & ~ & \multicolumn{5}{c}{\textit{Decision Trees (DT)}} \\ \cline{3-6} \cline{8-11} \cline{13-17}
     ~     &  & LinR  & LinR+CS & LinR+US & LinR+OS &  & LogR  & LogR+CS & LogR+US & LogR+OS &  & CDT   & DKMDT   & HDDT  & EAT   & ARDT  \\ \hline
Birds-s-thrush   &  & 0.48 (6)    & 0.46 (8)    & 0.30 (13) & 0.47 (7)    &  & 0.37 (12)   & 0.45 (9)   & 0.40 (11)  & 0.42 (10)  &  & 0.50 (3)    & 0.49 (4)    & 0.48 (5)    & 0.52 (2)   & \textbf{0.53} (1)   \\
Birds-kinglet    &  & 0.27 (3)    & 0.23 (8.5)  & 0.11 (12) & 0.23 (8.5)  &  & 0.19 (11)   & 0.27 (5)   & 0.22 (10)  & 0.27 (4)   &  & 0.24 (7)    & 0.00 (13)   & 0.25 (6)    & 0.32 (2)   & \textbf{0.38} (1)   \\
Birds-nighthawk  &  & \textbf{0.48} (1)    & 0.36 (2.5)  & 0.10 (12) & 0.36 (2.5)  &  & 0.16 (11)   & 0.33 (4)   & 0.23 (10)  & 0.29 (7)   &  & 0.28 (8)    & 0.08 (13)   & 0.24 (9)    & 0.33 (5)   & 0.32 (6)   \\
Letter-A         &  & 0.06 (12)   & 0.63 (7.5)  & 0.64 (6)  & 0.63 (7.5)  &  & 0.88 (5)    & 0.55 (11)  & 0.58 (10)  & 0.59 (9)   &  & 0.93 (3)    & 0.00 (13)   & \textbf{0.94} (1)    & 0.93 (4)   & 0.94 (2)   \\
Letter-B         &  & 0.00 (12.5) & 0.29 (9)    & 0.29 (8)  & 0.29 (10)   &  & 0.32 (7)    & 0.27 (11)  & 0.37 (6)   & 0.37 (5)   &  & 0.79 (3)    & 0.00 (12.5) & 0.80 (2)    & 0.76 (4)   & \textbf{0.83} (1)   \\
Letter-C         &  & 0.00 (12.5) & 0.44 (8)    & 0.43 (9)  & 0.44 (7)    &  & 0.47 (5)    & 0.30 (11)  & 0.41 (10)  & 0.46 (6)   &  & \textbf{0.87} (1)    & 0.00 (12.5) & 0.87 (3)    & 0.82 (4)   & 0.87 (2)   \\
Letter-D         &  & 0.00 (12.5) & 0.30 (8)    & 0.30 (10) & 0.30 (9)    &  & 0.54 (5)    & 0.30 (11)  & 0.40 (7)   & 0.42 (6)   &  & 0.81 (3)    & 0.00 (12.5) & 0.82 (2)    & 0.76 (4)   & \textbf{0.83} (1)   \\
Glass-containers &  & 0.13 (12.5) & 0.57 (5.5)  & 0.52 (9)  & 0.57 (5.5)  &  & 0.13 (12.5) & 0.43 (11)  & 0.52 (8)   & 0.62 (3)   &  & 0.53 (7)    & \textbf{0.68} (1)    & 0.61 (4)    & 0.52 (10)  & 0.66 (2)   \\
Glass-tableware  &  & 0.05 (13)   & 0.49 (7.5)  & 0.39 (9)  & 0.49 (7.5)  &  & 0.15 (12)   & 0.30 (11)  & 0.39 (10)  & 0.59 (5)   &  & 0.75 (4)    & 0.78 (2)    & 0.78 (2)    & 0.55 (6)   & 0.78 (2)   \\
Yeast-vac-vs-nuc &  & 0.00 (12)   & 0.32 (3.5)  & 0.30 (6)  & 0.32 (3.5)  &  & 0.00 (12)   & 0.17 (10)  & 0.28 (9)   & 0.28 (8)   &  & 0.28 (7)    & 0.00 (12)   & 0.37 (2)    & 0.32 (5)   & \textbf{0.38} (1)   \\
Yeast-me2        &  & 0.00 (12)   & 0.27 (7.5)  & 0.27 (4)  & 0.27 (7.5)  &  & 0.00 (12)   & 0.07 (10)  & 0.27 (6)   & 0.27 (5)   &  & 0.31 (2)    & 0.00 (12)   & \textbf{0.32} (1)    & 0.26 (9)   & 0.29 (3)   \\
Yeast-me1        &  & 0.00 (12.5) & 0.43 (7.5)  & 0.41 (9)  & 0.43 (7.5)  &  & 0.00 (12.5) & 0.06 (11)  & 0.39 (10)  & 0.47 (6)   &  & 0.65 (3)    & 0.67 (2)    & 0.61 (5)    & 0.62 (4)   & \textbf{0.69} (1)   \\
Yeast-exc        &  & 0.00 (12)   & 0.26 (6)    & 0.18 (9)  & 0.26 (7)    &  & 0.00 (12)   & 0.05 (10)  & 0.20 (8)   & 0.27 (5)   &  & 0.34 (4)    & 0.00 (12)   & 0.46 (2)    & 0.37 (3)   & \textbf{0.55} (1)   \\
Ecoli-om         &  & 0.53 (10)   & \textbf{0.78} (1.5)  & 0.73 (6)  & \textbf{0.78} (1.5)  &  & 0.00 (13)   & 0.12 (12)  & 0.46 (11)  & 0.70 (8)   &  & 0.76 (4)    & 0.57 (9)    & 0.75 (5)    & 0.73 (7)   & 0.77 (3)   \\
Newthyroid1      &  & 0.70 (13)   & 0.88 (8.5)  & 0.86 (10) & 0.88 (8.5)  &  & 0.94 (3)    & \textbf{0.96} (1.5) & 0.93 (4)   & \textbf{0.96} (1.5) &  & 0.85 (11)   & 0.93 (6)    & 0.93 (6)    & 0.84 (12)  & 0.93 (6)   \\
Newthyroid2      &  & 0.72 (13)   & 0.91 (8.5)  & 0.91 (11) & 0.91 (8.5)  &  & 0.96 (5)    & 0.97 (3)   & 0.93 (6)   & 0.96 (4)   &  & 0.89 (12)   & \textbf{0.97} (1.5)  & 0.91 (10)   & 0.92 (7)   & \textbf{0.97} (1.5) \\
Pageblocks0      &  & 0.60 (10)   & 0.60 (12)   & 0.60 (9)  & 0.60 (11)   &  & 0.65 (5)    & 0.63 (7)   & 0.64 (6)   & 0.62 (8)   &  & 0.83 (2)    & 0.00 (13)   & 0.83 (3)    & 0.83 (4)   & \textbf{0.83} (1)   \\
Led7digit1       &  & 0.00 (13)   & 0.50 (10.5) & 0.50 (12) & 0.50 (10.5) &  & 0.67 (6)    & 0.72 (5)   & 0.57 (8)   & 0.53 (9)   &  & 0.76 (4)    & 0.60 (7)    & 0.76 (2)    & 0.76 (2)   & 0.76 (2)   \\
\hline
    \end{tabular}}
    \caption{The experimental results in terms of f1-score (FSCORE). Numbers in parentheses show the relative ranking of the method on each dataset.}
    \label{table:fscore}

    \resizebox{1\textwidth}{!}{
    \begin{tabular}{l| c cccc c cccc c ccccc}
    \multicolumn{1}{c}{\multirow{2}{*}{ACC}}    & ~ & \multicolumn{4}{c}{\textit{Linear Regression (LinR)}}    & ~ & \multicolumn{4}{c}{\textit{Logistic Regression (LogR)}}    & ~ & \multicolumn{5}{c}{\textit{Decision Trees (DT)}} \\ \cline{3-6} \cline{8-11} \cline{13-17}
                      & ~ & LinR                     & LinR+CS      & LinR+US      & LinR+OS      & ~ & LogR                  & LogR+CS      & LogR+US      & LogR+OS      & ~ & CDT             & DKMDT        & HDDT      & EAT        & ARDT      \\ \hline
Birds-s-thrush   &  & 0.83 (7)    & 0.79 (9)    & 0.57 (13) & 0.80 (8)    &  & 0.85 (2.5)  & 0.74 (10)  & 0.70 (12)  & 0.71 (11)  &  & 0.84 (6)    & 0.85 (2.5)  & 0.85 (4.5)  & 0.85 (4.5) & \textbf{0.86} (1)   \\
Birds-kinglet    &  & 0.89 (7)    & 0.83 (8.5)  & 0.54 (13) & 0.83 (8.5)  &  & 0.94 (2)    & 0.79 (11)  & 0.72 (12)  & 0.79 (10)  &  & 0.92 (6)    & \textbf{0.94} (1)    & 0.92 (4)    & 0.92 (5)   & 0.93 (3)   \\
Birds-nighthawk  &  & 0.96 (2)    & 0.93 (8.5)  & 0.66 (13) & 0.93 (8.5)  &  & 0.95 (4)    & 0.87 (10)  & 0.78 (12)  & 0.83 (11)  &  & 0.94 (7)    & \textbf{0.96} (1)    & 0.95 (5)    & 0.95 (6)   & 0.95 (3)   \\
Letter-A         &  & 0.96 (6)    & 0.96 (9.5)  & 0.96 (8)  & 0.96 (9.5)  &  & 0.99 (5)    & 0.94 (13)  & 0.95 (12)  & 0.95 (11)  &  & 0.99 (3)    & 0.96 (7)    & \textbf{0.99} (1)    & 0.99 (4)   & 0.99 (2)   \\
Letter-B         &  & 0.96 (6.5)  & 0.82 (11)   & 0.82 (10) & 0.82 (12)   &  & 0.96 (5)    & 0.79 (13)  & 0.88 (9)   & 0.88 (8)   &  & 0.98 (3)    & 0.96 (6.5)  & 0.99 (2)    & 0.98 (4)   & \textbf{0.99} (1)   \\
Letter-C         &  & 0.96 (6.5)  & 0.91 (10)   & 0.91 (11) & 0.91 (9)    &  & 0.97 (5)    & 0.85 (13)  & 0.90 (12)  & 0.92 (8)   &  & \textbf{0.99} (1)    & 0.96 (6.5)  & 0.99 (3)    & 0.99 (4)   & 0.99 (2)   \\
Letter-D         &  & 0.96 (6.5)  & 0.82 (11)   & 0.82 (13) & 0.82 (12)   &  & 0.97 (5)    & 0.82 (10)  & 0.89 (9)   & 0.90 (8)   &  & 0.98 (3)    & 0.96 (6.5)  & 0.99 (2)    & 0.98 (4)   & \textbf{0.99} (1)   \\
Glass-containers &  & 0.94 (5)    & 0.90 (9.5)  & 0.87 (12) & 0.90 (9.5)  &  & 0.93 (7)    & 0.84 (13)  & 0.88 (11)  & 0.92 (8)   &  & 0.96 (2)    & 0.95 (4)    & \textbf{0.96} (1)    & 0.93 (6)   & 0.96 (3)   \\
Glass-tableware  &  & 0.96 (7)    & 0.88 (9.5)  & 0.83 (11) & 0.88 (9.5)  &  & 0.96 (5)    & 0.80 (13)  & 0.81 (12)  & 0.92 (8)   &  & 0.97 (4)    & 0.98 (2)    & 0.98 (2)    & 0.96 (6)   & 0.98 (2)   \\
Yeast-vac-vs-nuc &  & 0.94 (2)    & 0.79 (8.5)  & 0.75 (11) & 0.79 (8.5)  &  & 0.94 (2)    & 0.36 (13)  & 0.70 (12)  & 0.76 (10)  &  & 0.92 (5)    & 0.94 (2)    & 0.91 (6)    & 0.90 (7)   & 0.93 (4)   \\
Yeast-me2        &  & 0.97 (2)    & 0.86 (8.5)  & 0.84 (11) & 0.86 (8.5)  &  & 0.97 (2)    & 0.06 (13)  & 0.84 (12)  & 0.85 (10)  &  & 0.95 (4)    & 0.97 (2)    & 0.95 (6)    & 0.95 (7)   & 0.95 (5)   \\
Yeast-me1        &  & 0.97 (6.5)  & 0.92 (9.5)  & 0.91 (11) & 0.92 (9.5)  &  & 0.97 (6.5)  & 0.05 (13)  & 0.90 (12)  & 0.93 (8)   &  & 0.98 (2)    & 0.98 (4)    & 0.98 (3)    & 0.98 (5)   & \textbf{0.98} (1)   \\
Yeast-exc        &  & 0.98 (3)    & 0.88 (9)    & 0.83 (12) & 0.88 (10)   &  & 0.98 (3)    & 0.03 (13)  & 0.84 (11)  & 0.89 (8)   &  & 0.96 (7)    & 0.98 (3)    & 0.97 (5)    & 0.97 (6)   & \textbf{0.98} (1)   \\
Ecoli-om         &  & 0.97 (6)    & 0.96 (7.5)  & 0.94 (10) & 0.96 (7.5)  &  & 0.94 (11)   & 0.14 (13)  & 0.85 (12)  & 0.94 (9)   &  & 0.97 (3)    & 0.97 (5)    & 0.98 (2)    & 0.97 (4)   & \textbf{0.98} (1)   \\
Newthyroid1      &  & 0.93 (13)   & 0.97 (8.5)  & 0.96 (10) & 0.97 (8.5)  &  & 0.99 (3)    & \textbf{0.99} (1)   & 0.97 (7)   & 0.99 (2)   &  & 0.94 (11)   & 0.98 (5)    & 0.98 (5)    & 0.94 (12)  & 0.98 (5)   \\
Newthyroid2      &  & 0.94 (13)   & 0.97 (10)   & 0.97 (10) & 0.97 (10)   &  & 0.99 (4.5)  & 0.99 (3)   & 0.97 (8)   & 0.99 (4.5) &  & 0.97 (12)   & \textbf{0.99} (1.5)  & 0.98 (7)    & 0.98 (6)   & \textbf{0.99} (1.5) \\
Pageblocks0      &  & 0.94 (5)    & 0.89 (10)   & 0.89 (8)  & 0.89 (9)    &  & 0.94 (6)    & 0.89 (11)  & 0.89 (12)  & 0.88 (13)  &  & 0.97 (2)    & 0.90 (7)    & 0.97 (3)    & 0.96 (4)   & \textbf{0.97} (1)   \\
Led7digit1       &  & 0.92 (8)    & 0.84 (11.5) & 0.83 (13) & 0.84 (11.5) &  & 0.96 (5)    & 0.94 (7)   & 0.87 (9)   & 0.86 (10)  &  & 0.96 (4)    & 0.95 (6)    & 0.96 (2)    & 0.96 (2)   & 0.96 (2)   \\
 \hline
    \end{tabular}}\\
    \caption{The experimental results in terms of accuracy (ACC). Numbers in parentheses show the relative ranking of the method on each dataset.}
    \label{table:acc}

\end{table*}

\begin{figure}[t]
	\centering
	\subfigure[F1-score (FSCORE)]{\label{fig:fscore}\includegraphics[width=.5\textwidth]{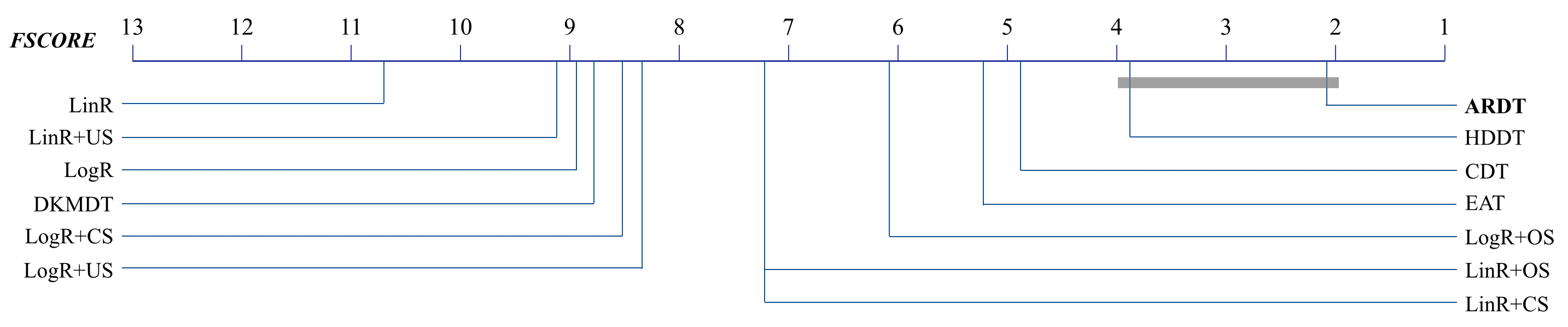}}
	\quad
	\subfigure[Accuracy (ACC)]{\label{fig:acc}\includegraphics[width=.5\textwidth]{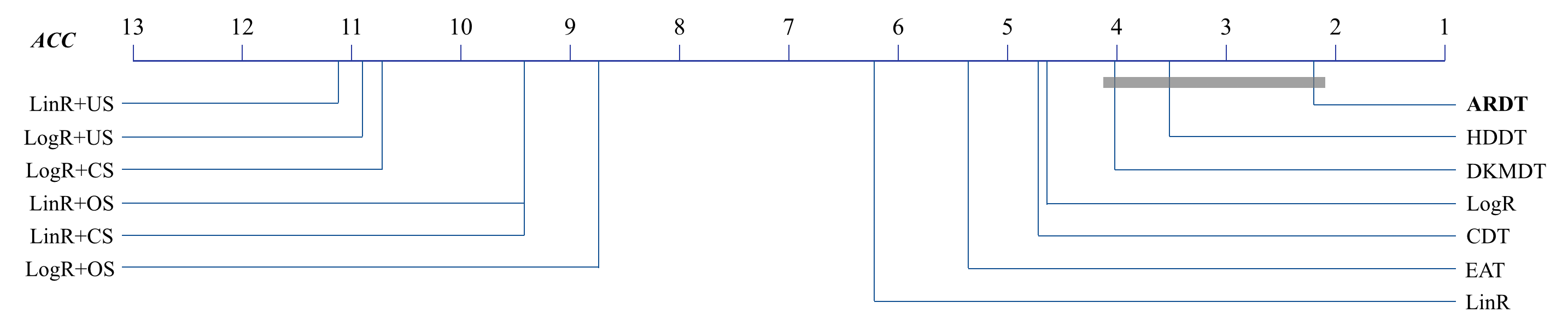}}
	\caption{Performance of the compared methods in terms of average rank using the Friedman test followed by Holm's step-down procedure at $\alpha=0.05$. The methods which are statistically equivalent to ARDT are connected with a gray line.}
	\label{fig:examples}
\end{figure}

\vspace{.5em}
\noindent
\textbf{Results:} \quadtab
Figures \ref{fig:fscore} and \ref{fig:acc} show the average rank (where 1 is best and 13 is worst) of the methods across all the datasets, in terms of FSCORE and ACC. All results are obtained using the Friedman test followed by Holm's step-down procedure with a 0.05 significance level \cite{Demsar:2006,Holm:1979}. We also report the detailed breakdown of the performance in tables \ref{table:fscore} and \ref{table:acc}. On each dataset, we perform ten-fold cross validation. The numbers in parentheses indicate the relative rank of the methods on each dataset. The best result on each dataset is shown in bold face.

In terms of FSCORE (Figure \ref{fig:fscore} and Table \ref{table:fscore}), our ARDT method produces the most preferable results. It outperforms all the other methods on nine datasets, and manages relatively high ranks on the rest datasets. This signifies that our method is able to improve the sensitivity ($\frac{TP}{TP+FN}$) while it maintains a low FP (that is, high precision ($\frac{TP}{TP+FP}$)). Based on the Friedman test, HDDT results in statistically equivalent results to our method, while CDT and EAT also produce competitive results.
On the other hand, although the conventional approaches (LinR+CS, LinR+US, LinR+OS, LogR+CS, LogR+US, and LogR+OS) show improvements over their base methods (LinR and LogR), their results are not as good as our method.

In terms of ACC (Figure \ref{fig:acc} and Table \ref{table:acc}), our ARDT method also performs the best. ARDT outperforms all the other methods on eight datasets and is evaluated as the best methods with HDDT and DKMDT by the Friedman test. However, we would like to point out that DKMDT is not a reliable method for our test datasets, because it fails to produce consistent FSCOREs which tells us DKMDT produces rather biased classification models. Similarly, the conventional approaches (LinR+CS, LinR+US, LinR+OS, LogR+CS, LogR+US, and LogR+OS) turn out decreasing ACC, compared to that of their base methods (LinR and LogR). This demonstrates that the conventional approaches are sacrificing a large number of negative (majority class) instances for a relatively smaller improvement in positive (minority) class. On the contrary, our ARDT method does not show such a tendency but results in reliable outputs both in terms of FSCORE and ACC.


\begin{figure}
\begin{adjustwidth}{-.07in}{-.07in}
\centering
\includegraphics[width=0.525\textwidth]{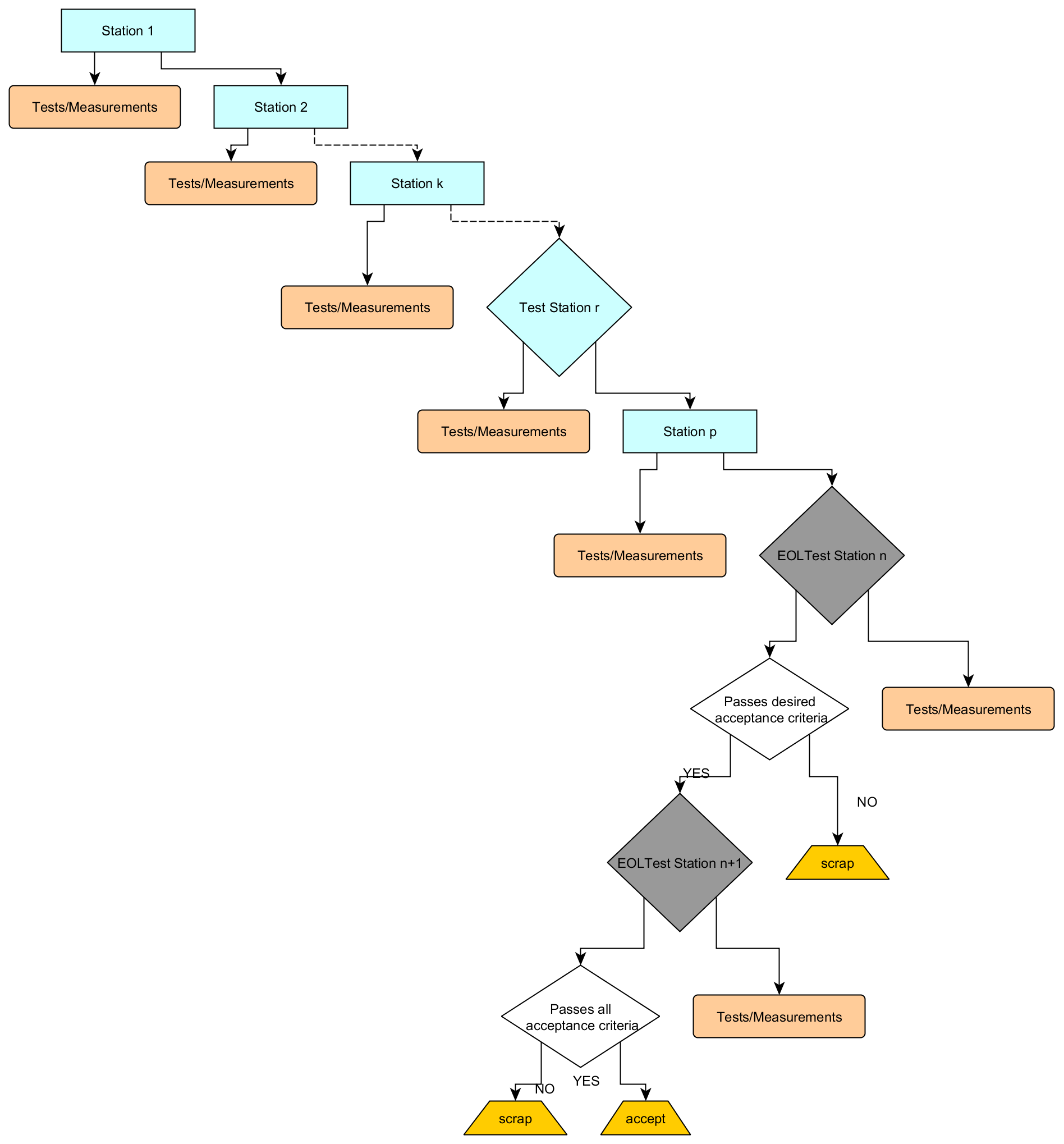}
\caption{A schematic representation of an assembly line}
\label{fig:AssemblyLine}
\end{adjustwidth}
\end{figure}

\subsection{Application to Manufacturing}
\label{subsec:exp2}
We apply the proposed method of thresholding to the manufacturing domain.  Our investigations focus on the production lines in manufacturing plants. 
Typically, an assembly line is associated with  multiple stations where different operations take place. In every station, several measurements are taken for each product instance up to that point.
Different components are added to an unfinished product in different  \emph{production stations} of an assembly line.

 An illustration of an assembly line is shown in Figure \ref{fig:AssemblyLine}. We have represented a production station by a rectangle.  
In the figure \emph{station 1, 2, k and p} in blue rectangles depict the production stations. 

At the end of an assembly line, there is usually a series of special testing stations inspecting the quality of the final finished product. These testing stations are called \emph{end-of-line (EOL)} testing stations.  In Figure \ref{fig:AssemblyLine},  a test station  is represented by a  rhombus. The EOL testing stations are shown in gray.

 If a product does not meet the required quality criteria, it is usually rejected or scrapped. A rejected product is called a \emph{scrap} or \emph{bad part} and an accepted product is called a \emph{good part}.   In an advanced manufacturing plant usually the amount of bad parts produced is very little as compared to good parts.

The information that is gathered and used in our study is from the measurements in an assembly line and the end of line tests.  The objective is to determine if scrap can be detected beforehand and what conditions leading to scrap. In this work, the product under investigation is a pump. Each instance of the pump  that is produced in this manufacturing process is called a part.

\vspace{.5em}
\noindent
\textbf{Description of the dataset used:} \quadtab
The data comprises of 16 factors or variables  and information for 5K parts manufactured within a period of 2 months with daily scrap rate fluctuating between $6-16\%$.

\vspace{.5em}
\noindent
\textbf{Methods and Metrics:} \quadtab
We evaluate the 6 outstanding methods (in FSCORE) from Section \ref{subsec:exp1}. Namely, we test ARDT, HDDT, CDT, EAT, LogR+OS, and LinR+OS on our dataset obtained from the manufacturing assembly line, and compare FSCORE and ACC of the results.

\begin{figure}
\centering
\includegraphics[width=.5\textwidth]{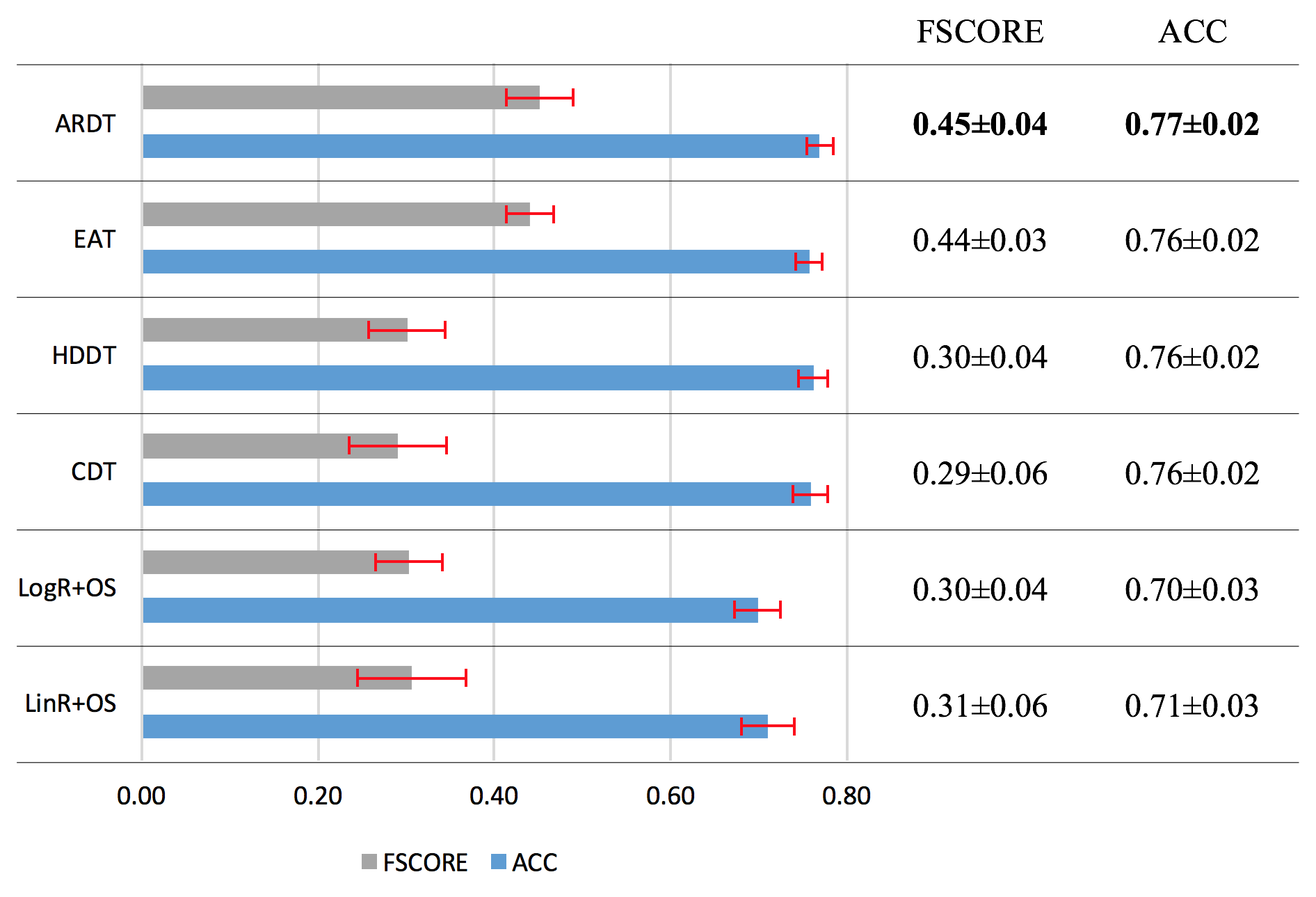}
\caption{Comparison of the results on the manufacturing assembly line data in terms of f1-score (FSCORE) and accuracy (ACC).}
\label{fig:exp2}
\end{figure}

\vspace{.5em}
\noindent
\textbf{Results:}\quadtab
Figure \ref{fig:exp2} shows the outlier (scrap) detection results in terms of FSCORE and ACC. All results are obtained from ten-fold cross validation. The numbers in bold face indicate the best result.

Based on the results, our ARDT method clearly outperforms the rest of the methods in terms of both FSCORE and ACC. 
This signifies that ARDT not only improves the accuracy in identifying the positive (scrap) class, but also maintains a good (overall) detection accuracy.
Interestingly, although statistically equivalent, ARDT shows even higher FSCORE and ACC than EAT, which builds and classifies using an ensemble of multiple R{\'e}nyi decision trees. 
We attribute this to the adaptive decision branches of ARDT that make the model as robust and precise as an ensemble model.
On the other hand, although HDDT and CDT produce competitive ACC, their low FSCORE (high FN or FP) makes the methods less preferable on our dataset.

Notice that all the decision tree models (ARDT, EAT, HDDT, CDT) show higher ACC than two of the over-sampled linear models (LogR+OS, LinR+OS). 
One possible explanation is that the data has a non-linear boundary between good and scrap parts, which could be captured by neither LogR+OS nor LinR+OS.

\vspace{.5em}
To summarize, through the empirical evaluation study, we tested and compared ARDT with other methods designed to address the class imbalance problem. 
Our observations strongly support the effectiveness of our ARDT method and its adaptive splitting criterion in solving the classification problem with class imbalance.
Our case study with the application to the manufacturing domain also confirms the capability of ARDT in addressing the scrap detection problem.
Our method has shown that it can effectually identify the outlying bad parts in a collection of the assembly line data.

\section{Conclusion}

In this paper, we formalized the concept of thresholding  and proposed a novel approach to exploit thresholding to improve classification in imbalanced datasets. 
We defined the concept of thresholding for linear classifiers. 
 With the aid of thresholding, we showed a principled methodology of addressing class imbalance for linear classifiers. We also demonstrated that thresholding is an implicit assumption for many approaches to deal with class imbalance. We then extended this paradigm beyond linear classification to develop a novel decision tree building method. 
Our approach incorporates thresholding with decision tree learning by devising a new splitting criterion that changes adaptively according to the underlying class distribution. Although we adopt the same R{\'e}nyi entropy as the existing methods, our method is different in that we decide the R{\'e}nyi parameter $\alpha$ according to the class distribution at each decision node. 
Our experiments on 18 publicly available imbalanced datasets showed that our proposed method is more accurate and robust than the compared methods in terms of both precision and sensitivity. 

By formulating the outlier detection problem as a classification problem where the outliers comprise of the rarer class, the proposed method can be used for outlier detection.  Taking the manufacturing domain as an example, we demonstrated the extensive applicability of this method in real-life scenarios.  In an advanced manufacturing process, where the scrap rate is very low, we showed that our method can be used to identify the outlying scraps with greater accuracy than the current state-of-the art methods.

Future work includes applying the novel ARDT method to other real-life use cases and demonstrating the concept of thresholding in other types of classifiers.

\section{Acknowledgments}
This work has been done while the primary author was at Robert Bosch LLC. The authors would like to thank Dr. Hyung-bo Shim for his insightful comments.

%
\bibliographystyle{abbrv}
\bibliography{ARDTOutlier_a}  
%
%

\end{document}